\documentclass{article}
\usepackage{ijcai20}

\usepackage{times}
\usepackage{soul}
\usepackage{url}
\usepackage[hidelinks]{hyperref}
\usepackage[utf8]{inputenc}
\usepackage[small]{caption}
\usepackage{graphicx}
\usepackage{amsmath}
\usepackage{amsthm}
\usepackage{amssymb}
\usepackage{booktabs}
\usepackage{algorithm}
\usepackage{algorithmic}
\usepackage{bm}
\usepackage{epstopdf}
\theoremstyle{plain} 
\theoremstyle{plain} 
\theoremstyle{plain} 
\theoremstyle{plain} 
\theoremstyle{plain} 
\theoremstyle{plain} 
\theoremstyle{plain} 
\theoremstyle{plain}

\title{Convolutional Graph-Tensor Net for Graph Data Completion}

\iftrue
\author{
Xiao-Yang Liu$^{1}$\and
Ming Zhu$^{2*}$
\affiliations
$^1$Columbia University\\
$^2$Shenzhen Institutes of Advanced Technology, Chinese Academy of Sciences \thanks{Corresponding authors.} 
\emails
xl2427@columbia.edu,
zhumingpassional@gmail.com
}
\fi

\begin{document}

\maketitle

\begin{abstract}

Graph data completion is a fundamentally important issue as data generally has a graph structure, e.g., social networks, recommendation systems, and the Internet of Things. We consider a graph where each node has a data matrix, represented as a \textit{graph-tensor} by stacking the data matrices in the third dimension. In this paper, we propose a \textit{Convolutional Graph-Tensor Net} (\textit{Conv GT-Net}) for the graph data completion problem, which uses deep neural networks to learn the general transform of graph-tensors. The experimental results on the ego-Facebook data sets show that the proposed \textit{Conv GT-Net} achieves significant improvements on both completion accuracy (50\% higher) and completion speed (3.6x $\sim$ 8.1x faster) over the existing algorithms.

\end{abstract}


\section{Introduction}
\begin{figure}[t]
\includegraphics[scale=0.26]{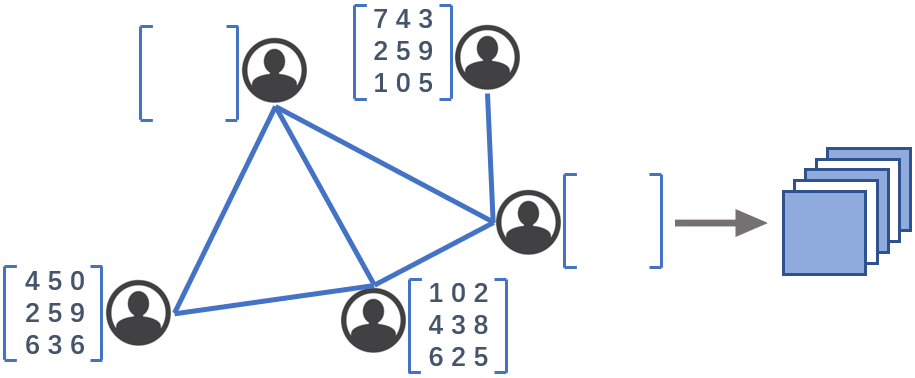}
\caption{Example of user data with graph structure. A graph-tensor can be obtained by stacking user profile matrices in the third dimension. }
\label{FGraph_Topology}
\end{figure}

Data with graph structures are common in real-world applications, e.g., user profiles in social networks, user-item matrices in recommendation systems, and sensory data in the Internet of Things \cite{liu2017ls,liu2016tensor}. Fig. \ref{FGraph_Topology} illustrates an example of user data with a graph structure in social networks or recommendation systems. Each user has a data matrix, and a \textit{graph-tensor} is obtained by stacking the data matrices in the third dimension. Due to the limitations of the data collection/measurement process \cite{liu2017ls}, it is common that only a subset of data matrices are observed while the other data matrices are totally unobserved. The key problem is how to complete a graph-tensor with missing data matrices. 

Several existing works are applied to the graph data completion problem. Tensor Alt-Min \cite{alt_min} and tensor nuclear-norm minimization with the alternative direction method of multipliers (TNN-ADMM) \cite{TNN_ADMM} are proposed based on the low-rank structure of the transform-based tensor model, which can be applied to the graph-tensor completion problem. \cite{Convolutional_Imputation} propose an iterative imputation algorithm for the graph-tensor completion problem, however, it is inadequate due to hundreds of iterations and the time-consuming shrinkage processes. The major challenge of the graph-tensor completion problem is to recover the missing data matrices by exploiting the graph topology.


\section{Graph-Tensor Model and Problem Formulation}
\begin{figure*}[t]
\includegraphics[scale=0.365]{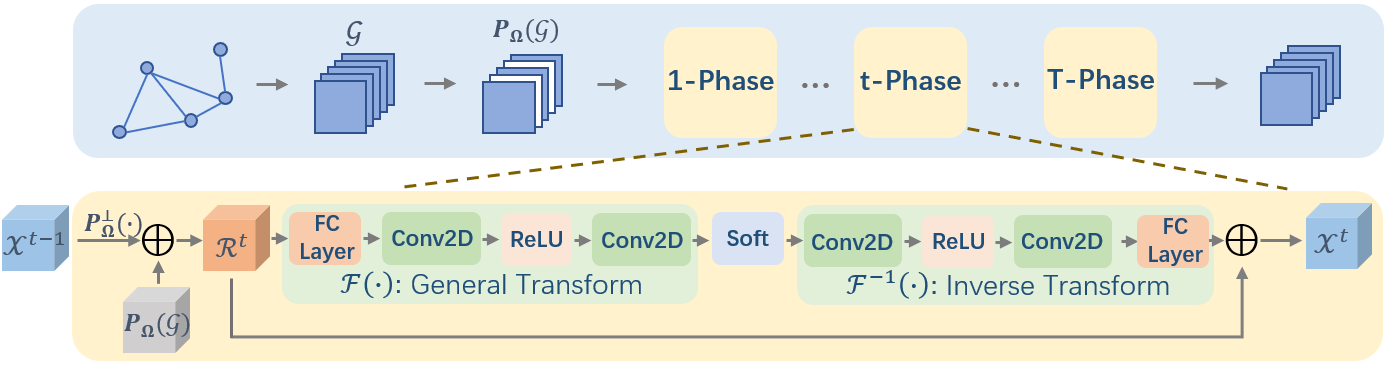}
\caption{The structure of \textit{Conv GT-Net}. The upper part is the data flow in \textit{Conv GT-Net}, containing $T$ phases. The bottom part is the detailed structure of one phase. Gray arrows denote the data flow.}
\label{FNet_Structure}
\end{figure*}

We represent a third-order tensor as $\mathcal{X} \in \mathbb{R}^{n_1 \times n_2 \times n_3}$, where the $(i,j)$-th tube is $\mathcal{X}(i,j,:)$ and the $k$-th frontal slice is $\mathcal{X}(:,:,k)$, or $\mathcal{X}^{(k)}$ for simplicity. We use $[n]$ to denote the set $\{1,...,n\}$.  The Frobenius norm of a third-order tensor $\mathcal{X}$ is defined as $\|\mathcal{X}\|_F = \sqrt{\sum_{i=1}^{n_1}\sum_{j=1}^{n_2}\sum_{k=1}^{n_3}|\mathcal{X}(i,j,k)|^2}$. 

\subsection{Graph-Tensor Model}

Consider an undirected graph $G=(V, E)$, where $V$ denotes a set of nodes with $|V| = n_3$, and $E$ denotes a set of edges. We assume that each node has a data matrix $\mathcal{X}^{(k)}\in\mathbb{R}^{n_1 \times n_2}$. After stacking the data matrices along the third dimension, we obtain a \textit{graph-tensor} $\mathcal{X}\in\mathbb{R}^{n_1 \times n_2 \times n_3}$.

Let $\bm{A}\in\mathbb{R}^{n_3 \times n_3}$ be the symmetric adjacency matrix of graph $G$, where the diagonal entries are zeros. The corresponding non-negative diagonal degree matrix $\bm{D}\in\mathbb{R}^{n_3 \times n_3}$ is $\bm{D}(k,k)=\sum_{s=1}^{n_3}\bm{A}(k,s),~\text{for}~k\in[n_3]$.
The graph Laplacian matrix of $G$ is a real symmetric matrix $\bm{L}=\bm{D}-\bm{A}\in\mathbb{R}^{n_3 \times n_3}$, and the normalized graph Laplacian matrix of $G$ is $\bm{L}_n = \bm{I} - \bm{D}^{-\frac{1}{2}}\bm{A}\bm{D}^{-\frac{1}{2}}\in\mathbb{R}^{n_3 \times n_3}$, where $\bm{D}^{-\frac{1}{2}}$ is calculated in an element-wise manner on the diagonal entries $\bm{D}(k,k)$ for $k\in[n_3]$.

Let the eigenvalue decomposition of $\bm{L}$ be $\bm{L}=\bm{Q}\bm{E}\bm{Q}^{H}$, where $\bm{Q}\in\mathbb{R}^{n_3 \times n_3}$ consists of the eigenvectors, $\bm{E}\in\mathbb{R}^{n_3 \times n_3}$ is a diagonal matrix with the corresponding eigenvalues, and $H$ is the Hermitian transpose. The graph transform matrix $\bm{U}\in\mathbb{R}^{n_3 \times n_3}$ is a unitary matrix defined as $\bm{UL}=\bm{EU}$, and $\bm{U}=\bm{Q}^{H}$. 

Any linear transform can be the transform $\mathcal{L}$ in \cite{Liu2017Fourth-order}. In this paper, we use the graph Fourier transform as the transform $\mathcal{L}$, which is performed by the matrix $\bm{U}$. Let $\widetilde{\mathcal{X}}$ denote the graph-tensor in the spectral domain for $\mathcal{X}$. The graph Fourier transform and the inverse graph Fourier transform can be defined as follows: 
\begin{align}
    \label{EGraph_Fourier_Transform}
    \widetilde{\mathcal{X}}^{(k)} &= \sum_{s\in[n_3]}\bm{U}(k,s)\mathcal{X}^{(s)},~\text{for}~k\in[n_3],\\
    \label{EGraph_Inverse_Fourier_Transform}
    ~\mathcal{X}^{(k)} &= \sum_{s\in[n_3]}\bm{U}^{-1}(k,s)\widetilde{\mathcal{X}}^{(s)},~\text{for}~k\in[n_3],
\end{align}
where $\bm{U}^{-1}$ is the inverse matrix of $\bm{U}$.

\subsection{Problem Formulation}

We assume that the graph-tensor in the spectral domian is low rank, which has been supported in many research works, such as the spectral space singular value distribution \cite{Convolutional_Imputation}. We assume that a subset of data matrices of a graph-tensor are observed. For example, for a graph-tensor $\mathcal{G}\in\mathbb{R}^{n_1 \times n_2 \times n_3}$ with $n_3$ nodes, the observed data matrices are $P_\Omega(\mathcal{G})$, where $P_\Omega:\mathbb{R}^{n_1\times n_2\times n_3}\rightarrow \mathbb{R}^{n_1\times n_2\times n_3}$ is the projection of the tensor to a partial observation by only retaining entries in the index set $\Omega \in [n_3]$. Let $\mathcal{X}$ be a graph-tensor which has the same partial observation as $P_\Omega(\mathcal{G})$. For a graph-tensor $\mathcal{X} \in\mathbb{R}^{n_1 \times n_2 \times n_3}$, the $\mathcal{L}$-SVD \cite{Liu2017Fourth-order} is $\mathcal{X}=\mathcal{U} \bullet \mathcal{S}\bullet\mathcal{V}^{\dagger}$. We define its graph-tensor nuclear-norm as $\|\mathcal{X}\|_{\text{gTNN}} = \langle \mathcal{S},\mathcal{I}\rangle=\sum_{i=1}^{r}|\widetilde{\mathcal{S}}(i,i,1)|$, where $r=\text{rank}_{\mathcal{L}}(\mathcal{X})$. 

To recover the data matrices of the unobserved nodes, the graph-tensor completion problem is formulated as:
\begin{equation}
    \label{EProblem}
    \begin{split}
    &\min_{\widetilde{\mathcal{X}}} \|\widetilde{\mathcal{X}}\|_{\text{gTNN}}, \\
    &\text{s.t.}~ \mathcal{P}_{\Omega}(\mathcal{X}) = \mathcal{P}_{\Omega}(\mathcal{G}).
    \end{split}
\end{equation}

Let $\mathcal{P}_{\Omega}^{\perp}$ be the complement of $\mathcal{P}_{\Omega}$, and $\bm{A} = \bm{V_1} \Sigma \bm{V_2}^H$ be the singular value decomposition. Let $\text{singular-soft}(\cdot)$ denote the diagonal soft-thresholding operator on the singular value vector of the matrix: $\text{singular-soft}(\bm{A},\lambda) = \bm{V_1} (\Sigma - \lambda\bm{I})_+ \bm{V_2}^H$ with $(\cdot)_+$ keeping only positive values.

Problem~(\ref{EProblem}) can be solved via the iterative imputation algorithm in \cite{Convolutional_Imputation}. Its key steps are as follows for $t\ge 1$ and $k\in[n_3]$:
\begin{align}
    \label{EImputation_Projection}
    \mathcal{R}^{t}~~~~ &= ~ \mathcal{P}_{\Omega}(\mathcal{G}) + \mathcal{P}_{\Omega}^{\perp}(\mathcal{X}^{t-1}),\\
    \label{EImputation_Transform}
    \widetilde{\mathcal{R}}^{t(k)} &= ~ \sum_{s=1}^{n_3}\bm{U}(k,s)\mathcal{R}^{t(s)},\\
    \label{EImputation_Shrinkage}
    \mathcal{\widetilde{X}}^{t(k)} &= ~ \text{singular-soft}(\widetilde{\mathcal{R}}^{t(k)}, \bm{\lambda}(k)),\\
    \label{EImputation_Inverse}
    \mathcal{X}^{t(k)} &= ~ \sum_{s=1}^{n_3}\bm{U}^{-1}(k,s)\widetilde{\mathcal{X}}^{t(s)},
\end{align}
where $\mathcal{X}^{0} = \bm{0}$.

\section{Convolutional Graph-Tensor Net}
We propose a \textit{Convolutional Graph-Tensor Net} (\textit{Conv GT-Net}) for the graph data completion problem in this section. The basic idea is to map the imputation algorithm steps in (\ref{EImputation_Projection})-(\ref{EImputation_Inverse}) into deep neural networks with a fixed number of phases, where each phase corresponds to one iteration. We assume that there are $T$ phases in the \textit{Conv GT-Net} with the same parameters. Fig.~\ref{FNet_Structure} shows the structure of \textit{Conv GT-Net}. 

\textbf{1) Imputation.}
Step (\ref{EImputation_Projection}) aims to keep the observed nodes the same as their ground truth. We use the same operation in (\ref{EImputation_Projection}) as the imputation results in each phase of \textit{Conv GT-Net}.


\textbf{2) General Transform.}
Step (\ref{EImputation_Transform}) uses graph Fourier transform in (\ref{EGraph_Fourier_Transform}) to obtain the low-rank data in the spectral domain. Considering the great representation power of the convolutional neural networks, we adopt one fully connected (FC) layer, two 2D convolution layers and one ReLU activation layer as the general transform structure, denoted by $\mathcal{F}(\cdot) = \text{Conv2D}(\text{ReLU}(\text{Conv2D}(\text{FC}(\cdot))))$. For simplicity, we use $3\times 3$ filters in convolution layers. $\mathcal{F}(\cdot)$ aims to learn the general transform of the graph-tensor. The graph-tensor $\mathcal{R}^{t}$ in the spectral domain is denoted by $\mathcal{F}(\mathcal{R}^{t})$.

\begin{figure}[t]
\includegraphics[scale=0.48]{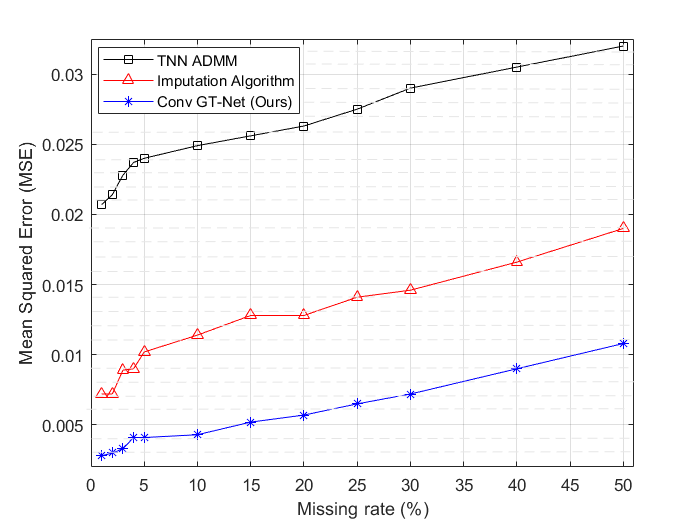}
\caption{Experimental results of the unweighted graph-tensor. Completion accuracy vs. missing rate.}
\label{FUnweighted_Completion_Accuracy}
\end{figure}

\textbf{3) Soft-Thresholding Operator.}
We replace (\ref{EImputation_Shrinkage}) by the soft-thresholding operator in \cite{soft-thresholding}. Unlike the diagonal soft-thresholding operator, the soft-thresholding operator is defined as $\text{soft}(a, \lambda) = \text{max}(|a|-\lambda,0)\cdot \text{sign}(\text{a})$, where $\text{sign}(\cdot)$ is the sign of $a$. It is an element-wise operator on each entry when applied to a matrix or tensor. Deep neural networks have great learning ability, so we set the soft-thresholding parameter set $\bm{\lambda}\in[T]$ to be learnable instead of using the pre-set ones in \cite{Convolutional_Imputation}. The graph-tensor after the soft-thresholding operator is denoted by $\text{soft}(\mathcal{F}(\mathcal{R}^{t}), \bm{\lambda})$.

\textbf{4) Inverse Transform.}
We introduce an inverse transform structure $\mathcal{F}^{-1}(\cdot)$ as the inversion of the general transform to replace step (\ref{EImputation_Inverse}), i.e. $\mathcal{F}^{-1}(\mathcal{F}(\mathcal{X})) = \mathcal{X}$. Specifically, $\mathcal{F}^{-1}(\cdot)$ adopts the symmetric structure of $\mathcal{F}(\cdot)$ and  $\mathcal{F}^{-1}(\cdot) = \text{FC}(\text{Conv2D}(\text{ReLU}(\text{Conv2D}(\cdot))))$. The filters in $\mathcal{F}^{-1}(\cdot)$ share the same size as those in $\mathcal{F}(\cdot)$. The graph-tensor after all the above operations is $\mathcal{F}^{-1}(\text{soft}(\mathcal{F}(\mathcal{R}^{t}), \bm{\lambda}))$.

\textbf{Shortcut Structure.}
Inspired by the residual network \cite{residual-net}, we add a shortcut structure before $\mathcal{X}^{t}$ to make the network more robust:
\begin{equation}
    \label{EXn}
    \mathcal{X}^{t} = \mathcal{R}^{t} + \mathcal{F}^{-1}(\text{soft}(\mathcal{F}(\mathcal{R}^{t}), \bm{\lambda}).
\end{equation}

\textbf{Loss Function.} 
Two items should be considered in \textit{Conv GT-Net}: the fidelity of the reconstructed data, and the accuracy of the inverse function. Loss function $\mathcal{L}$ is the weighted summation of two terms:
\begin{align}
    \label{ELoss}
    \mathcal{L} ~~~~~~&=~ \alpha\mathcal{L}_{\text{fidelity}} + \beta\mathcal{L}_{\text{inversion}},\\
    \label{EFidelity}
    \mathcal{L}_\text{fidelity} ~~&=~ \|P_{\Omega}^{\perp}(\mathcal{X}^{T}) - P_{\Omega}^{\perp}(\mathcal{G})\|_F^2, \\
    \label{EInversion}
    \mathcal{L}_\text{inversion} ~&=~ \frac{1}{T}\sum_{t=1}^{T}\|{\mathcal{F}^{-1}}(\mathcal{F}(\mathcal{X}^{t})) - \mathcal{X}^{t}\|^2_F.
\end{align}
Referring to \cite{ISTA-Net} and considering the case in our work, we set $\alpha = 1$ and $\beta = 0.0001$.

\begin{figure}[t]
\includegraphics[scale=0.48]{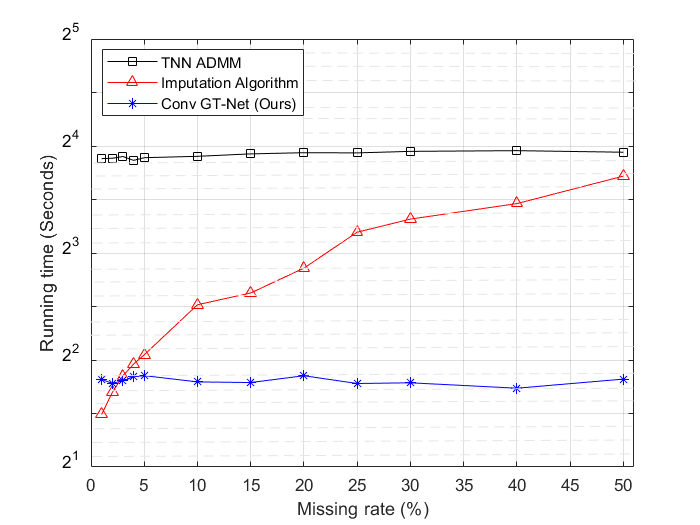}
\caption{Experimental results of the unweighted graph-tensor. Running time vs. missing rate.}
\label{FUnweighted_Completion_Time}
\end{figure}

\section{Performance Evaluation}

The comparison algorithms are TNN-ADMM \cite{TNN_ADMM} and imputation algorithm \cite{Convolutional_Imputation}, which do not need training and are implemented using MATLAB. \textit{Conv GT-Net} is trained in TensorFlow on a server with an NVIDIA Telsa V100 GPU (16GB device memory). For a fair comparison, we perform all the experiments on a laptop with an i7-8750H CPU (2.20GHz) and 16GB memory. The phase number $T$ is 10, the learning rate is 0.0001, and the running epoch is 500. Adam optimizer \cite{adam} is used to minimize the loss function. 

We take the graph topology of the ego-Facebook data sets from SNAP \cite{snapnets}, which contain an unweighted and undirected graph with 4,039 nodes and 88,234 edges. We pick 100 nodes (No. 896-995) and their corresponding edges to form a graph topology. To evaluate the performance in graph data with weighted edges, we randomly assign a weight (1-20) to each edge to form another graph topology. We use the feature data of the ego-Facebook data sets, select the nodes whose feature vector has 576 features and resize the feature vectors into $24 \times 24$ feature matrices. We regard the feature matrices as the data in the spectral domain and perform the inverse graph Fourier transform to get the graph-tensor in the time domain. Consequently, the graph-tensor has a size of $24 \times 24 \times 100$, and we select nodes and their feature data to form 32 graph-tensors as testing data and another 900 graph-tensors as training data. 

\begin{figure}[t]
\includegraphics[scale=0.48]{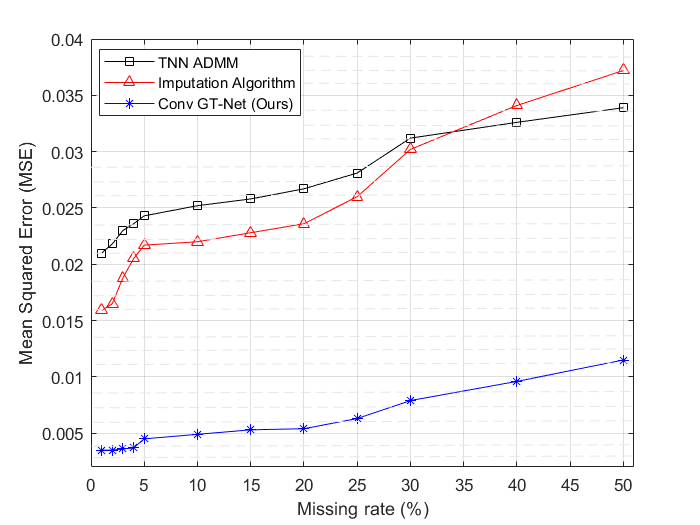}
\caption{Experimental results of the weighted graph-tensor. Completion accuracy vs. missing rate.}
\label{FWeighted_Completion_Accuracy}
\end{figure}

\begin{figure}
\includegraphics[scale=0.48]{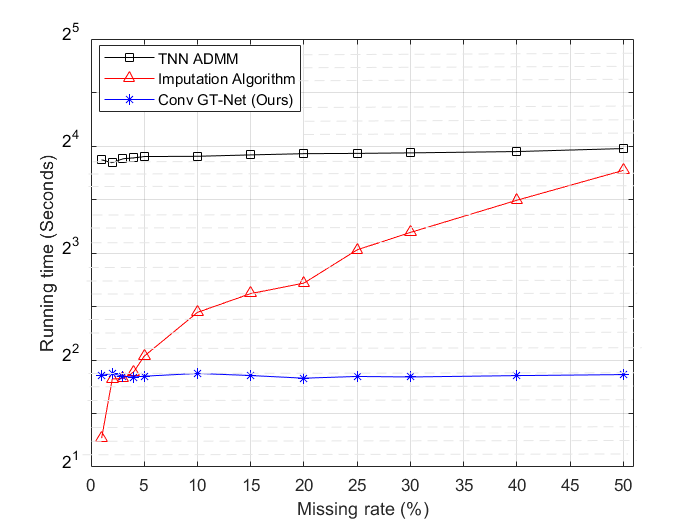}
\caption{Experimental results of the weighted graph-tensor. Running time vs. missing rate.}
\label{FWeighted_Completion_Time}
\end{figure}

We take mean squared error (MSE) and running time as metrics. Fig. \ref{FUnweighted_Completion_Accuracy} shows the completion accuracy of the unweighted graph-tensors. We observe that $\textit{Conv GT-Net}$ provides an average of 50\% higher completion accuracy than the imputation algorithm, let alone TNN-ADMM. This is mainly due to the strong low-rank assumption of the imputation algorithm since the real data is usually not strictly low-rank in the spectral domain. However, $\textit{Conv GT-Net}$ can learn a general transform for real data, which alleviates the dependency on the low-rank assumption. TNN-ADMM provides the lowest accuracy since it omits the graph topology. 

Fig. \ref{FUnweighted_Completion_Time} shows the running time of the unweighted graph-tensors. We observe that \textit{Conv GT-Net} runs about 3 to 8 times faster than the imputation algorithm and about 4 times faster than TNN-ADMM. It is worth noting that for different missing rates, \textit{Conv GT-Net} and TNN-ADMM keep a constant running speed, whereas the imputation algorithm \cite{Convolutional_Imputation} needs more running time as the missing rate increases. This is because $\textit{Conv GT-Net}$ adopts a fixed structure for a certain size of graph-tensors and TNN-ADMM omits the graph topology, while the imputation algorithm needs more iterations to converge as the missing rate increases.

Fig. \ref{FWeighted_Completion_Accuracy} and Fig. \ref{FWeighted_Completion_Time} show the completion accuracy and running time of the weighted graph-tensor, respectively. We notice that $\textit{Conv GT-Net}$ and TNN-ADMM are less influenced by the weight change of the graph while the imputation algorithm is more influenced. This is because the imputation algorithm performs graph Fourier transform which takes the weights information into consideration.

\vspace{-0.1in}

\section{Conclusion}

In this paper, we proposed \textit{Conv GT-Net} to solve the graph-tensor completion problem using deep neural networks. It learns a general transform of a graph-tensor. The experimental results show that \textit{Conv GT-Net} provides higher completion accuracy and runs faster than the other algorithms.

\vspace{-0.1in}
\section*{Acknowledgment}

Ming Zhu was supported by National Natural Science Foundations of China (Grant No. 61902387).

\bibliographystyle{ijcai20}

\end{document}